\documentclass{article}
\usepackage{spconf,amsmath,graphicx}
\usepackage{hyperref}
\usepackage{mathrsfs}
\usepackage{epsfig}
\usepackage{scrextend}
\usepackage{subfigure}
\usepackage{float}
\usepackage[]{algorithm2e}
\usepackage{stackengine}
\usepackage[export]{adjustbox}
\usepackage{booktabs}
\usepackage{tikz}
\usetikzlibrary{spy}

\usepackage[font=small]{caption}
\graphicspath{{IMAGES/}}

\title{Divergence-Based Adaptive Extreme Video Completion}

\name{Majed El Helou \qquad Ruofan Zhou \qquad Frank Schmutz \qquad Fabrice Guibert \qquad Sabine S{\"u}sstrunk}
\address{School of Computer and Communication Sciences, EPFL, Switzerland}

\begin{document}
\small
\maketitle

\begin{abstract}
Extreme image or video completion, where, for instance, we only retain 1\% of pixels in random locations, allows for very cheap sampling in terms of the required pre-processing. The consequence is, however, a reconstruction that is challenging for humans and inpainting algorithms alike. 

We propose an extension of a state-of-the-art extreme image completion algorithm to extreme video completion. We analyze a color-motion estimation approach based on color KL-divergence that is suitable for extremely sparse scenarios. Our algorithm leverages the estimate to adapt between its spatial and temporal filtering when reconstructing the sparse randomly-sampled video. We validate our results on $50$ publicly-available videos using reconstruction PSNR and mean opinion scores.
\end{abstract}

\begin{keywords}
Extreme completion, sparse color motion, extreme compression, video inpainting.
\end{keywords}

\section{Introduction} \label{sec:intro}
Image completion is a challenging task for which a rich literature exists. Different interpolation or inpainting methods, based on total variation, partial differential equations and matrix completion, or self-similarity, can be leveraged to reconstruct unknown pixel intensities~\cite{1,2,3,4,5,6,7,8,9}. They are, however, limited by their computational cost, as discussed in~\cite{achanta}. A simple way of extending these methods to video is by applying them frame by frame as mentioned in~\cite{4}. But the computational cost, already high per image, of techniques with comparable reconstruction performance to~\cite{achanta}, becomes too large over videos.
Another challenge is accounting for the additional temporal constraints of videos. A video completion technique must reconstruct the full frames, but also preserve temporal stability such that flickering does not become an issue~\cite{ilan2015survey}.

Most existing techniques for object removal~\cite{8}, inpainting~\cite{le2017motion}, or hole filling~\cite{liu2009video}, rely on the presence of at least a sufficient continuous portion of the video. A recent approach even leverages a combination of video compression and inpainting methods~\cite{jennifer}, but also relies on more significant amounts of available information. From the available portion of the video, relevant information can be extracted, such as a motion field~\cite{shiratori2006video}, to assist the completion. A more efficient version is developed~\cite{bokov2018100+}, but is also not valid under extremely sparse sampling, where the vast majority of pixels is unknown (e.g.: 99\% missing pixels).
\begin{figure}[t]
	\centering
	\subfigure[Reference $|$ 1\%-sampled frame]{
		\includegraphics[width=0.47\linewidth]{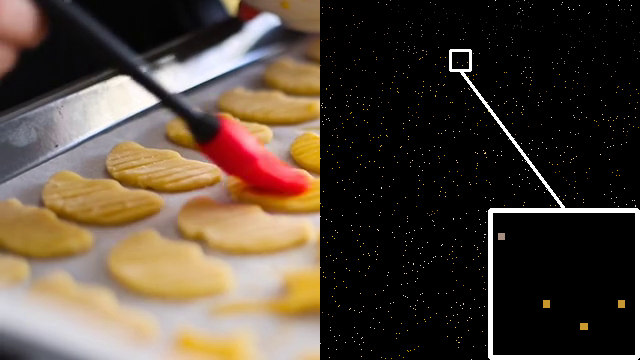}
	}
	\subfigure[EFAN2D (20.75 dB)]{
		\includegraphics[width=0.47\linewidth]{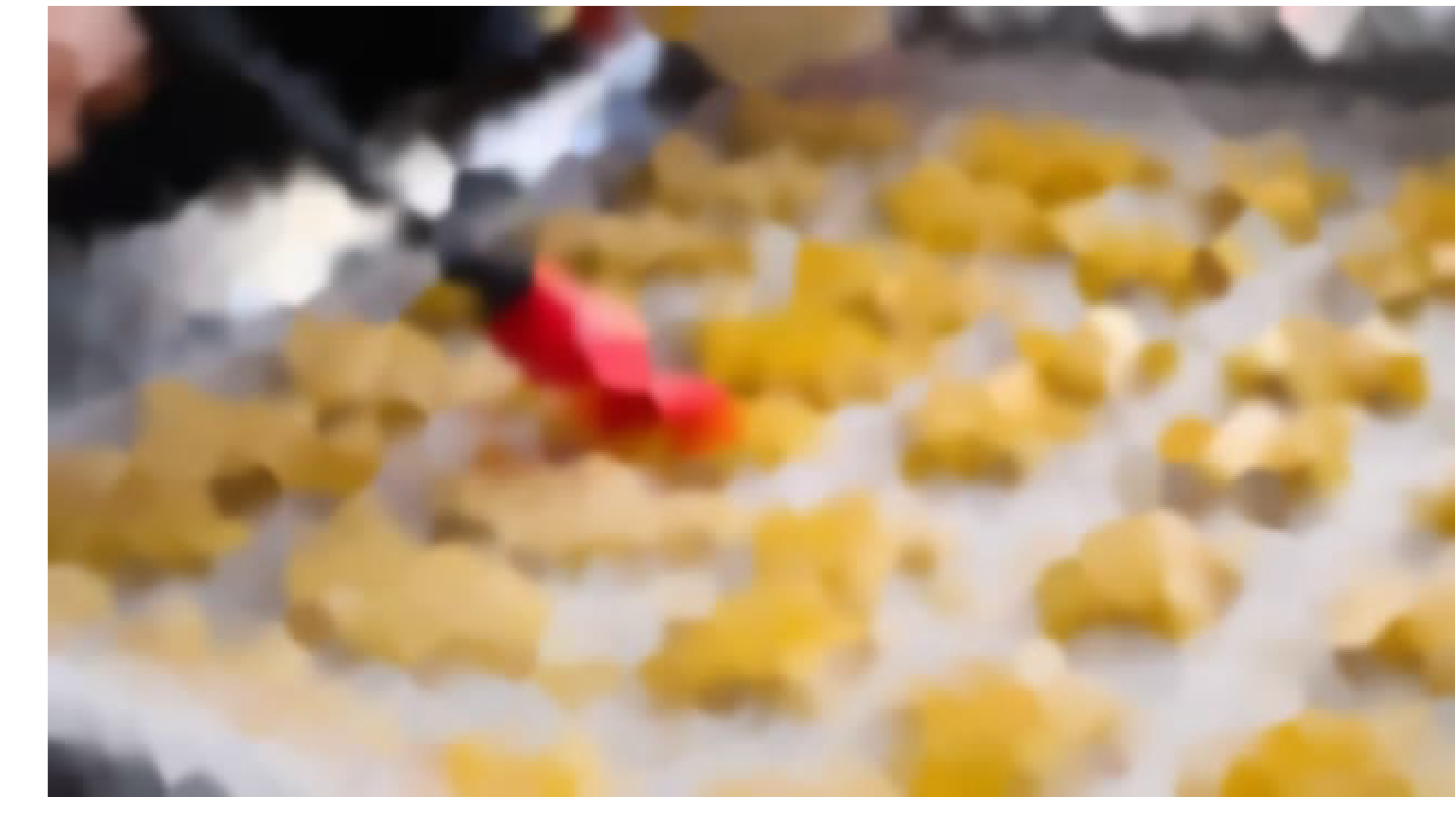}
	}
	\subfigure[EFAN3D (20.39 dB)]{
		\includegraphics[width=0.47\linewidth]{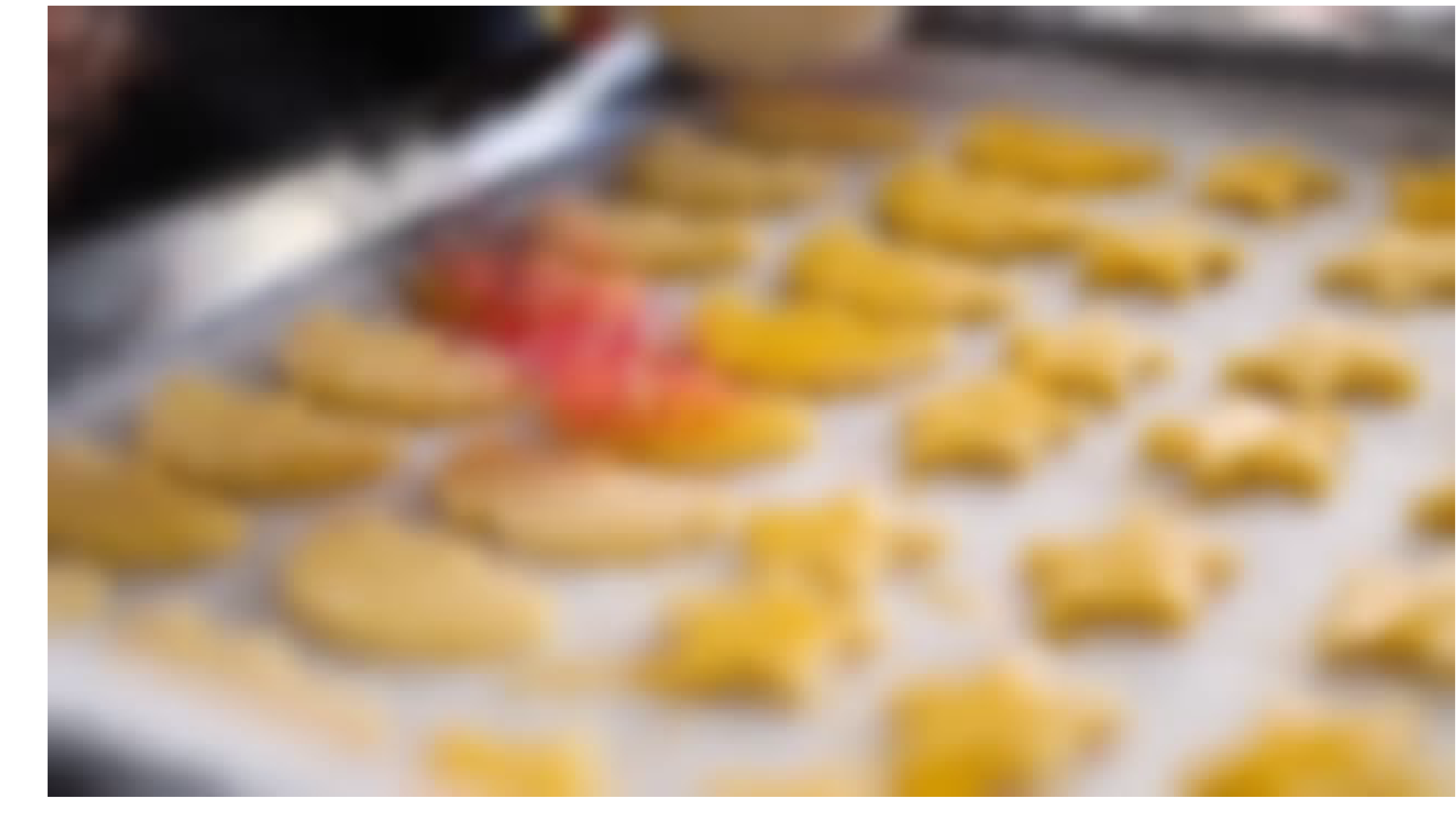}
	}
	\subfigure[ADEFAN (\textbf{22.02 dB})]{
		\includegraphics[width=0.47\linewidth]{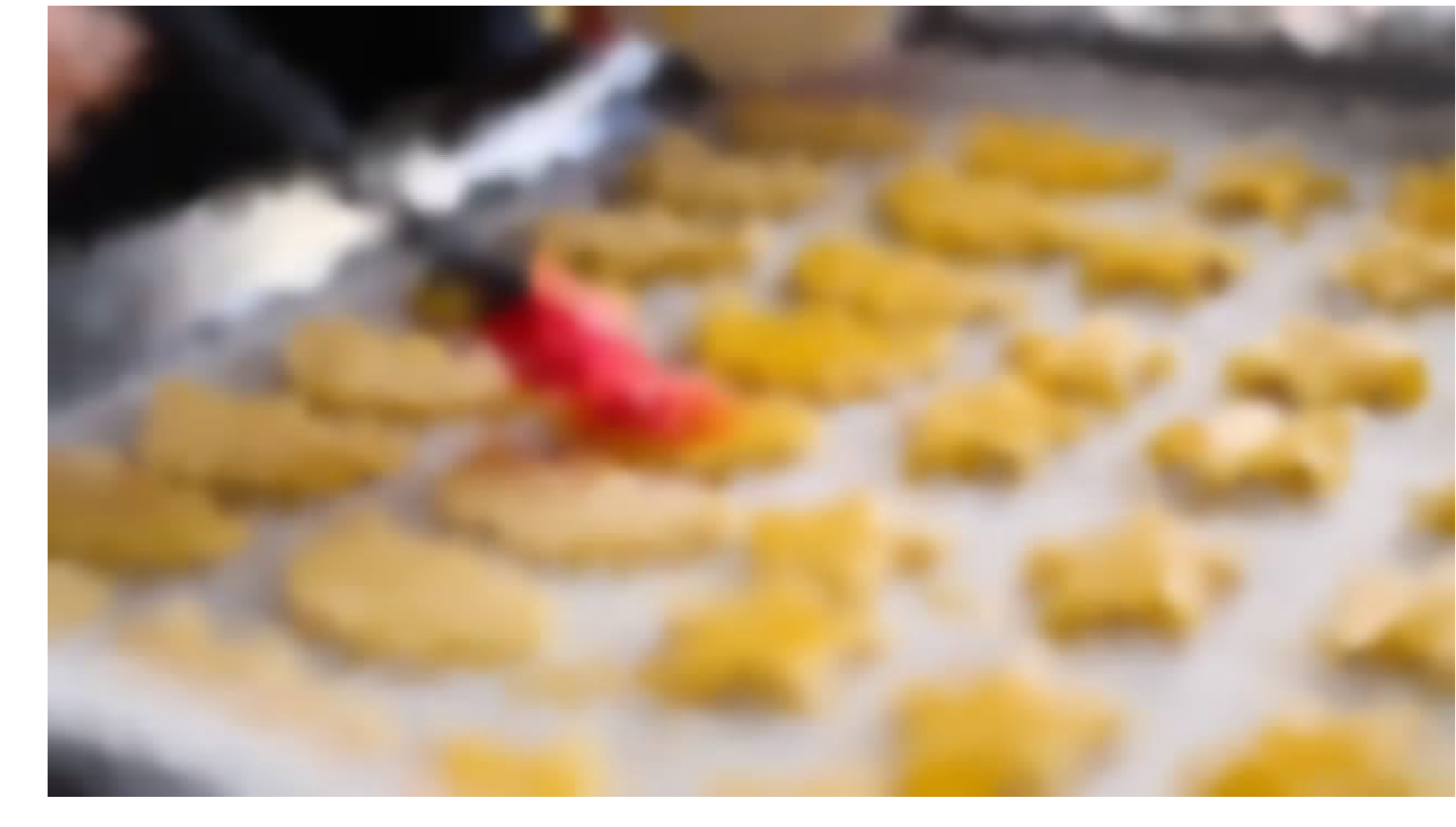}
	}
	\vspace{-0.1cm}
	\caption{The sample reference frame is reconstructed from 1\% randomly-sampled pixels (a), using different methods. (b) EFAN2D can track motion but causes disturbing video flickering. (c) EFAN3D achieves video stability at the cost of excessive temporal blurring effects. (d) Our adaptive ADEFAN method leverages the advantages of both (b) and (c). We achieve the best video reconstruction accuracy with minimal flickering. Best viewed on screen.}
	\label{fig:images}
\end{figure}

We propose an adaptive-depth extreme video completion method (ADEFAN), extending on the computationally-efficient extreme image completion method proposed by Achanta et al.~\cite{achanta}. The latter, Efficient Filtering by Adaptive Normalization (EFAN), is multiple orders of magnitude faster than its closest competitors. 
We first present a divergence-based approach for estimating color motion in extremely sparsely-sampled videos. We evaluate the accuracy of our sparse color motion estimation computed on 1\%-sampled videos, relative to the full-pixel frames. We then leverage this color motion estimation in our adaptive video completion method. 
Our completion filtering is carried out per color channel to account for potential aberration problems~\cite{el2018aam,mayer2018accurate}, but it could be applied across channels to make use of spectral correlation~\cite{el2017correlation}.
We consider frames with 1\% of pixels sampled randomly to test our completion methods on a public dataset of 50 videos. Our ADEFAN proposed method improves in terms of PSNR on the reference by $0.94dB$ on average.

\section{Filtering by Adaptive Normalization}\label{sec:EFAN2D3D}

\subsection{EFAN2D}\label{sec:E2D}
We begin with a brief overview of the extreme image completion method, EFAN~\cite{achanta}. EFAN performs filtering by adaptive normalization by summing up at each unknown pixel position $i$ the Gaussian-weighted contributions of known pixels $k$ in its neighborhood. The completed output $J[i]$ from a sparse image $I$ is
\begin{equation} \label{eq:equation1}
    J[i] = \frac{\sum_{k\in \mathcal{N}} G[i, k]I[k]}{ \sum_{k\in \mathcal{N}} G[i, k]  },
\end{equation}
where $G[i, k]$ is the filter defined as
\begin{equation}
    G[i, k] = e^{-0.5\frac{(i_x-k_x)^2 + (i_y-k_y)^2}{ \sigma^2 }},
\end{equation}
and where $\mathcal{N}$ is a completion neighborhood around pixel $i$, $(i_x,i_y)$ are pixel $i$'s coordinates, and $\sigma$ is a chosen parameter. We choose $\sigma$ = $\sqrt{\frac{1}{f * \pi}}$, where $f$ is the fraction of sampled pixels and $\mathcal{N}$ is a square of side length $2 * \left \lceil{0.5 + 3 * \sigma}\right \rceil + 5$, as in~\cite{achanta}.

A first extension to 3D completion (i.e., video completion, where the third dimension is time) is to apply EFAN frame by frame. We refer to this approach as EFAN2D, and use the same weights proposed in~\cite{achanta}. One advantage of this technique is that it can preserve motion throughout the video. However, as no temporal consistency is enforced from frame to frame, this results in pixel flickering in videos which is visually unappealing.
The flickering is particularly pronounced as the extreme sampling is random. This motivates the search for a time-aware approach that accounts for video stability.

\subsection{EFAN3D}\label{sec:E3D}
To overcome the aforementioned temporal instability of EFAN2D, we can leverage information from adjacent frames in the extreme completion. 
Neighboring frames contribute to the completion of a given frame by carrying the completion step across entire frame sets. To that end, Eq. (~\ref{eq:equation1}) is extended to take into account pixels $k$ from a set of adjacent frames. As distance in time is not equivalent to distance in space, we incorporate a time-specific exponential weight $\sigma_t$ and the expression becomes
\begin{equation} \label{eq:equation2}
    J[i] = \frac{\sum_{k\in \mathcal{N}} G[i, k]G_{t}[i, k]I[k]}{ \sum_{k\in \mathcal{N}} G[i, k]G_{t}[i, k]  },
\end{equation}
where $G_{t}[i, k]$ is the filter defined as
\begin{equation}
    G_{t}[i, k] = e^{-0.5\frac{(i_z-k_z)^2}{ \sigma_t^2 }},
\end{equation}
and where $\mathcal{N}$ is now a 3D completion neighborhood around pixel $i$, and $i_z$ is a temporal coordinate.
We choose the temporal weight $\sigma_t$ such that 99 frames contribute to the completion: the previous 49 frames, the current frame and the next 49 frames. It is set to $49/6$, as beyond $6\sigma_t$ the exponential weight becomes negligible.

EFAN3D does improve temporal consistency over EFAN2D, but at the cost of motion loss. More specifically, moving components are severely blurred out across frames. Results can be more visually appealing as flickering is absent, but temporal movement is prone to blurring and a lot of motion information is lost.

The trade-off in reconstruction performance between EFAN2D and EFAN3D thus depends on the amount of movement in the video. To obtain the best possible extreme video completion results, an approach more similar to EFAN2D should be applied on high-motion regions and rather more like EFAN3D on low-motion regions. The challenge, however, is two-fold. First, object motion in itself is not necessarily important, because if a uniform object is displaced over a fixed window it does not affect reconstruction. It is motion of texture and color that is most relevant. Second, the color motion estimation must be carried out under extreme conditions, i.e., 1\% pixel sampling, since the reconstruction algorithm cannot access the full video frames. We discuss our approach in the following section.

\begin{figure}[t!]
	\centering
	\subfigure[$80\times 80$ windows]{
		\includegraphics[width=0.46\linewidth,height=0.46\linewidth]{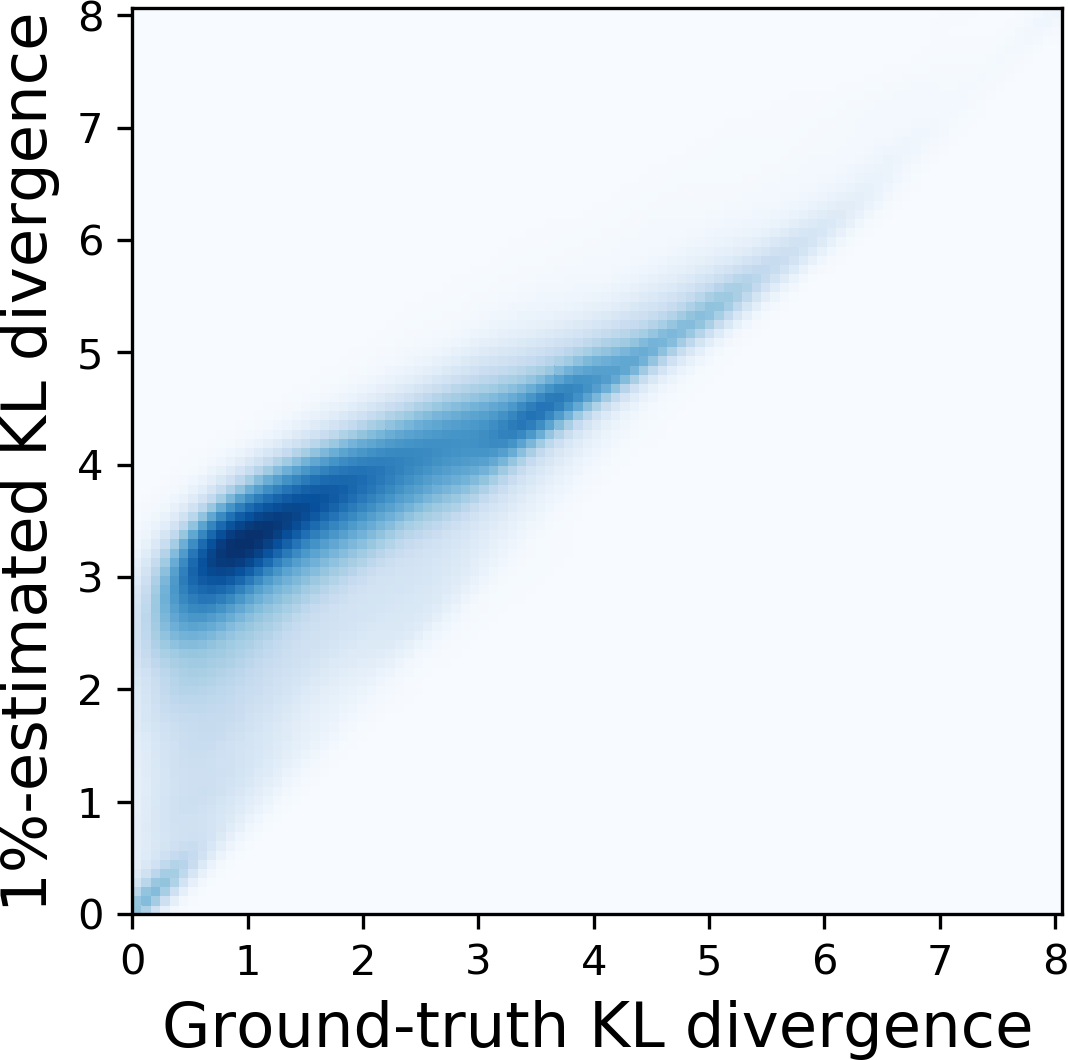}
	}
	\subfigure[$160\times 160$ windows]{
		\includegraphics[width=0.46\linewidth,height=0.46\linewidth]{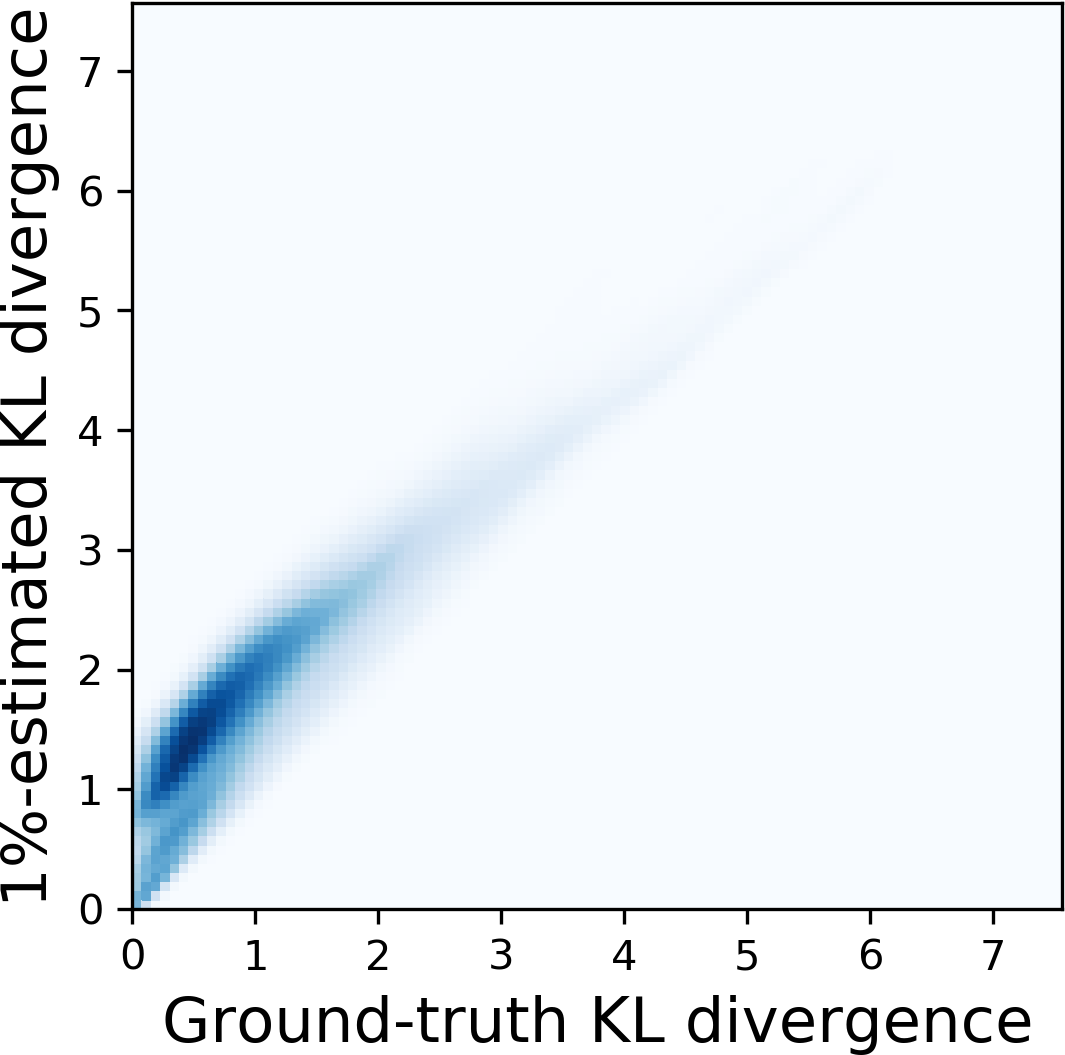}
	}
	\caption{Relation between our 1\%-based KL divergence estimates and full window KL divergence over different window sizes. The estimation becomes more accurate with larger windows.}
	\label{fig:window_size}
\end{figure}

\section{Adaptive extreme video completion}\label{sec:sparseKL}
We first present an approach for estimating color motion on extremely sparsely-sampled videos. We then develop an adaptive extreme video completion approach leveraging this motion estimation.

\subsection{Sparse color motion estimation}\label{sec:div_change}
Multiple approaches can be used to estimate color motion in videos, from simple color gradients to more complex texture evolution algorithms. Under extremely sparse and random pixel sampling, we are, however, restricted in terms of the applicable techniques. Gradient computation is, for instance, not possible between frames since the pixel locations are randomly sampled. Therefore, it is possible that no overlap exists between consecutive frames. Approximating gradients from closest neighbors comes at an exponential complexity cost and is thus not an option for videos. 

We thus propose to study the distribution of RGB color intensities between two frames. 
We denote by $ \textbf{p} $ the intensity distribution of frame $l$ and by $ \textbf{q} $ that of frame $l+m$. 
To estimate variation between these two frames, we compute the KL divergence~\cite{KL,bayesian} between $ \textbf{p} $ and $ \textbf{q} $. However, with extreme sampling it is likely to have, for some color $c$, $ \textbf{q}[c] > 0$ when $ \textbf{p}[c] = 0 $. To avoid the division by zero, we offset both distributions through the addition of a weighted uniform distribution $\textbf{u}[c]$
\begin{equation} \label{eq:equation3}
\begin{aligned}
    &\textbf{\~p}[c] = \alpha \textbf{p}[c] + (1-\alpha)\textbf{u}[c],
\end{aligned}
\end{equation}
where $\textbf{u}[c]= 1/256 $ for eight bit intensity values. We then compute the divergence on the offset distributions
\begin{equation} \label{eq:equation3b}
\begin{aligned}
    &div(\textbf{p,q}) = D_{KL}(\textbf{\~p} || \textbf{\~q}).
\end{aligned}
\end{equation}

We empirically find that $\alpha$ in the range [$10^{-5}$,$1-10^{-5}$] yield accurate divergence estimation. For the remainder of this paper, we choose $\alpha = 0.95$.
One shortcoming of KL divergence for color motion estimation is that it is oblivious to object rearrangement within the scene. Considering the case where frame $i+m$ is a mirrored version of frame $i$, divergence would not detect the in-frame motion. We address this shortcoming by processing small windows of the frames. Under extreme completion, however, we cannot use very small windows as pixels are sparse and a small window would contain too few pixels for an accurate KL divergence estimation. 
We compare KL divergence estimates between two 1\%-sampled video windows to the ground-truth divergence computed between the two fully-sampled windows for 3,000 frames taken from 50 different videos (60 frames per video), for different window sizes. The heat-map distributions are shown in Fig.~\ref{fig:window_size} for a window size of 80$\times$80 and 160$\times$160. 
The accuracy of divergence estimation increases with window size as seen in Fig.~\ref{fig:window_size}. For the larger window size, the correlation is higher between the 1\%-based KL estimates and the KL divergence computed on the full windows. We thus select the smallest possible window size that gives a KL estimation error below a predefined threshold to balance the trade-off between having small precise windows and accurate KL estimation. We adapt this window size to the sampling rate. We expand on this further in Sec.~\ref{sec:adefan}.

\begin{figure}[t!]
	\centering
	\subfigure{
		\includegraphics[width=.98\linewidth, trim={0 0 0 0}, clip]{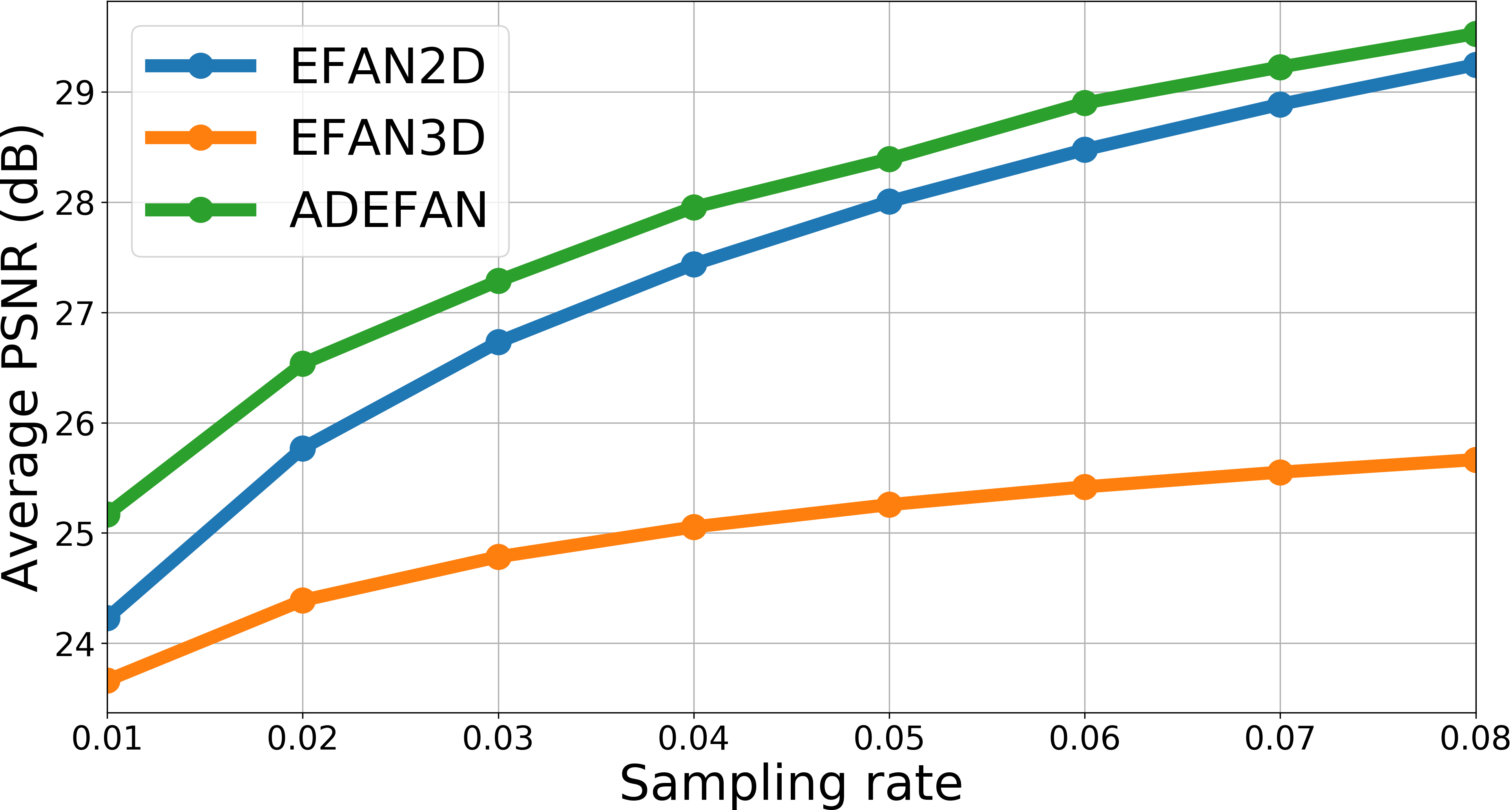}
	}
	\caption{Average PSNR results of each method for different extreme sampling rates (from 1\% to 8\% of pixels).}
	\label{fig:results_lineplot}
\end{figure}

\begin{table}[t]
    \centering
    \caption{Mean opinion scores for reconstruction and visual quality on a scale from 1 to 10, with 10 being perfect, gathered from 25 participants on 8 random videos (one for each sampling rate 1\% to 8\%).}
    \begin{tabular}{cccc}
        \toprule
        Opinion Score & EFAN2D & EFAN3D & ADEFAN \\ \hline
        {Reconstruction} & $\underline{5.53}$ & $3.72$ & $\textbf{6.52}$ \\
        {Visual Quality} & $\underline{5.05}$ & $4.12$ & $\textbf{6.34}$ \\
        \bottomrule
    \end{tabular}
    \label{tab:res_}
\end{table}

\begin{figure*}[t]
	\centering
	\subfigure{
		\includegraphics[width=0.98\linewidth]{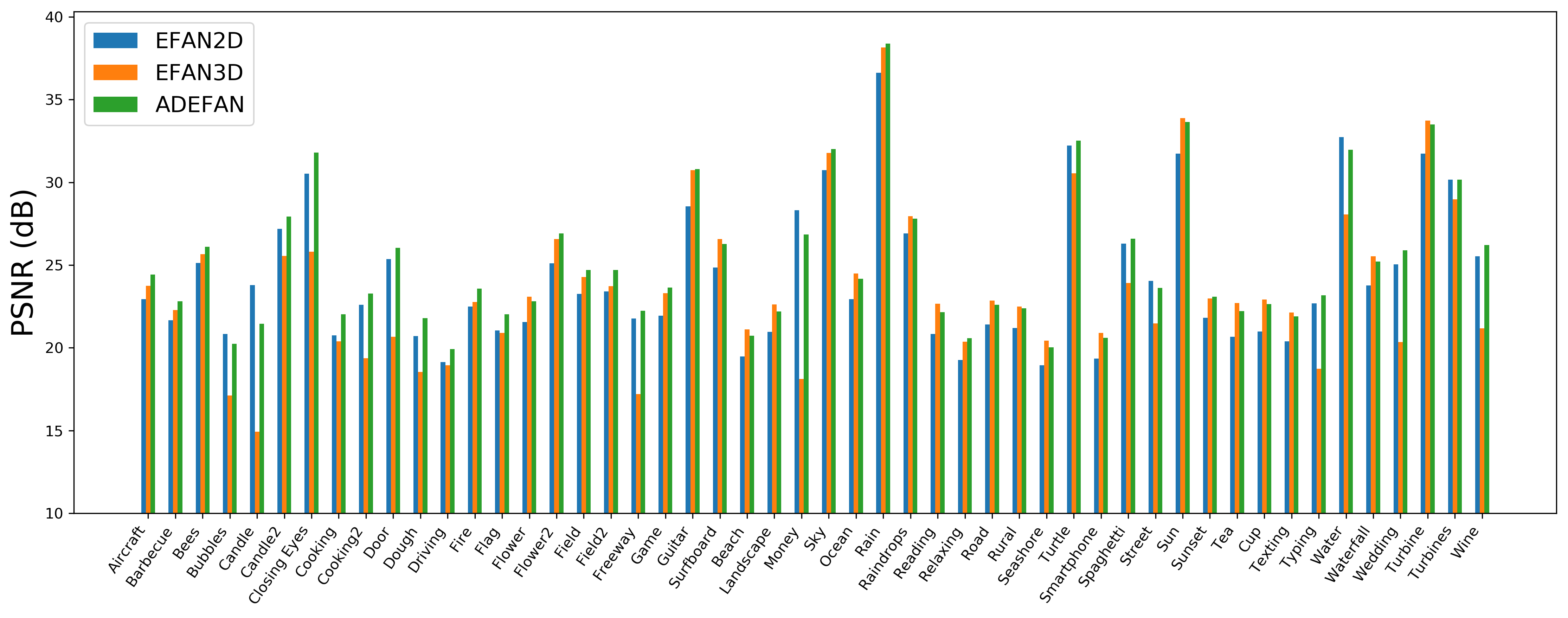}
	}
	\vspace{-.5cm}
	\caption{PSNR results for the three reconstruction methods on each video in the dataset, with a 1\% sampling rate.}
	\label{fig:results_barplot}
\end{figure*}

\subsection{Adaptive-Depth EFAN (ADEFAN)}
\label{sec:adefan}
We propose the adaptive-depth EFAN method, ADEFAN, to leverage the advantages of both EFAN2D and EFAN3D by relying on sparse color motion estimation. ADEFAN adapts the depth, in number of frames, across which the completion filtering is carried out, with a depth of 1 being the special case of EFAN2D. When less temporal color motion is estimated in a given window, more adjacent frames can be considered for the completion. Depth is computed independently per window.
We select the window size for each sampling rate based on the mean squared error between the ground-truth KL divergence and the KL divergence estimated from the sampled video. We empirically find that window sizes that have a mean squared error in the range $[0.15, 0.2]$ have a good compromise between small size and KL-estimation accuracy.

For a window in any given frame $l$ in the video, we estimate the KL divergence between it and the previous frames and also between it and the next frames. To estimate the divergence between frame $l$ and the next frames, we compute the divergence between frames $l$ and $l+1$, $l$ and $l+2$ and so on until the divergence between frames $l$ and $l+m+1$ is smaller than that between frames $l$ and $l+m$. We call $div_{next}$ the divergence between frames $l$ and $l+m$. The analogous $div_{prev}$ for previous frames is computed in the same way.
Forward depth is then defined as inversely proportional to the estimated divergence. To allow ADEFAN to flexibly vary between EFAN2D-like and EFAN3D-like completion, forward depth is given as a function of $div_{next}$ and the maximum number $fr_{max}$ of next frames we consider (which is the same as in EFAN3D)
\begin{equation} \label{eq:equation4}
    f(div_{next}, fr_{max}) = \left \lceil{ \frac{fr_{max}}{1 + \beta div_{next} }}\right \rceil.
\end{equation}
The addition of $1$ in the denominator avoids division by zero and ensures a maximal depth of $fr_{max}$, the ceiling ensures a minimal depth of 1, and $\beta$ is a parameter that controls the impact of the divergence. We empirically fix $\beta = 14$ for all of our experiments. Backward depth is computed in an analogous manner.

The time-depth $f$ of the filter is computed for every spatial window in the video frame. As we make the windows overlap for better accuracy, we need to combine all their results to reconstruct a frame. To avoid edge artifacts around a window, each of its pixels is given a Gaussian weight according to its distance to the center of the window. We use $\sigma = L/6$, where $L$ is the side length of the window, to compute the Gaussian weights. The final pixel value is the weighted average of its value in all the overlapping windows.

\newcommand{\insetA}[1]{
\begin{tikzpicture}[baseline=0em]
    \begin{scope}[spy using outlines=
          {magnification=2, size=.8cm, connect spies}]
        \node { \includegraphics[width=0.42\linewidth]{#1}};
        \spy [red] on (-0.4,0.0) in node [left] at (1.7,0.5);
    \end{scope}
\end{tikzpicture} 
}

\begin{figure}[t]
	\centering
	\subfigure[Reference video frame]{
		\insetA{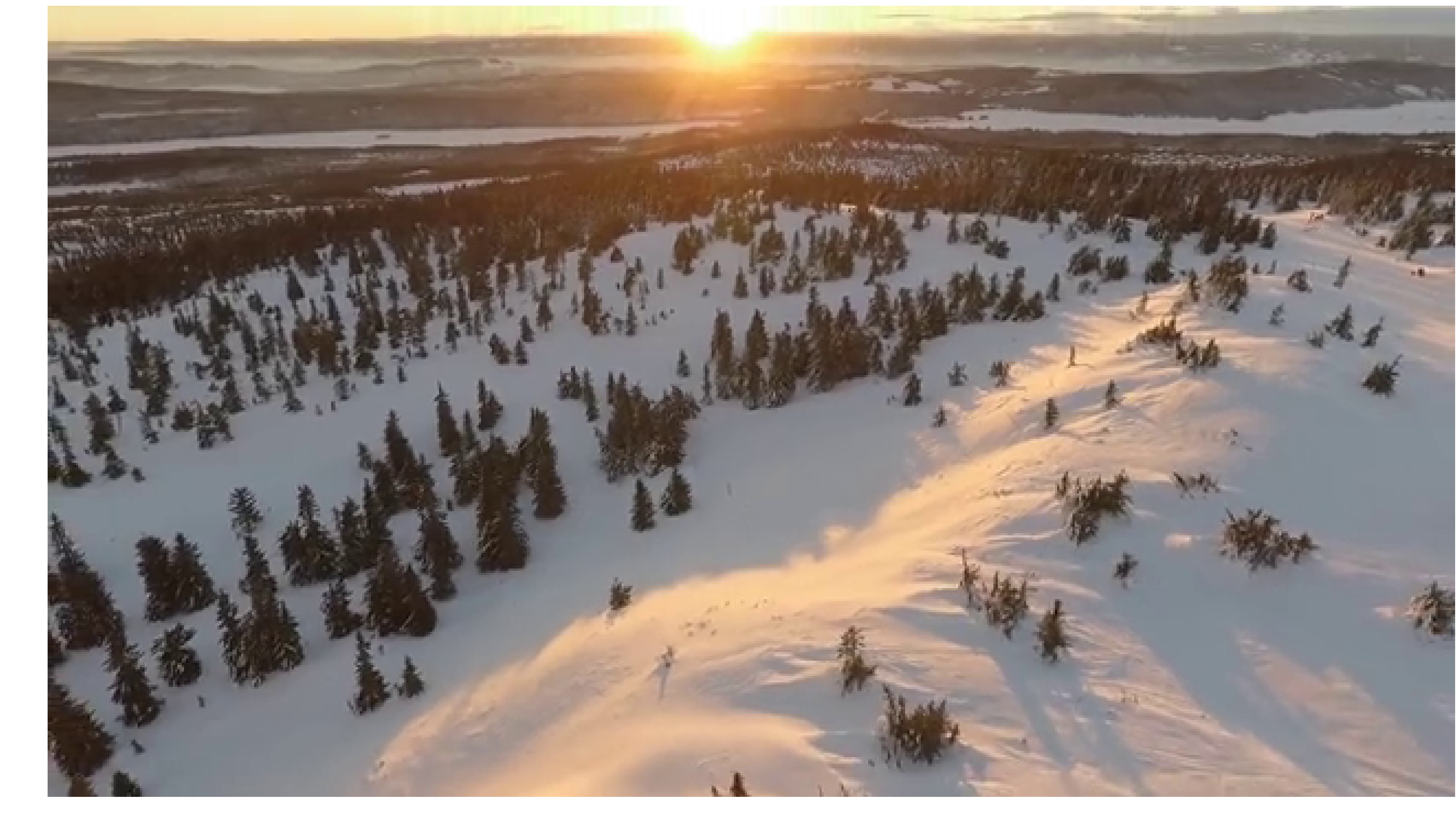}
	}
	\subfigure[EFAN2D (24.78 dB)]{
        \insetA{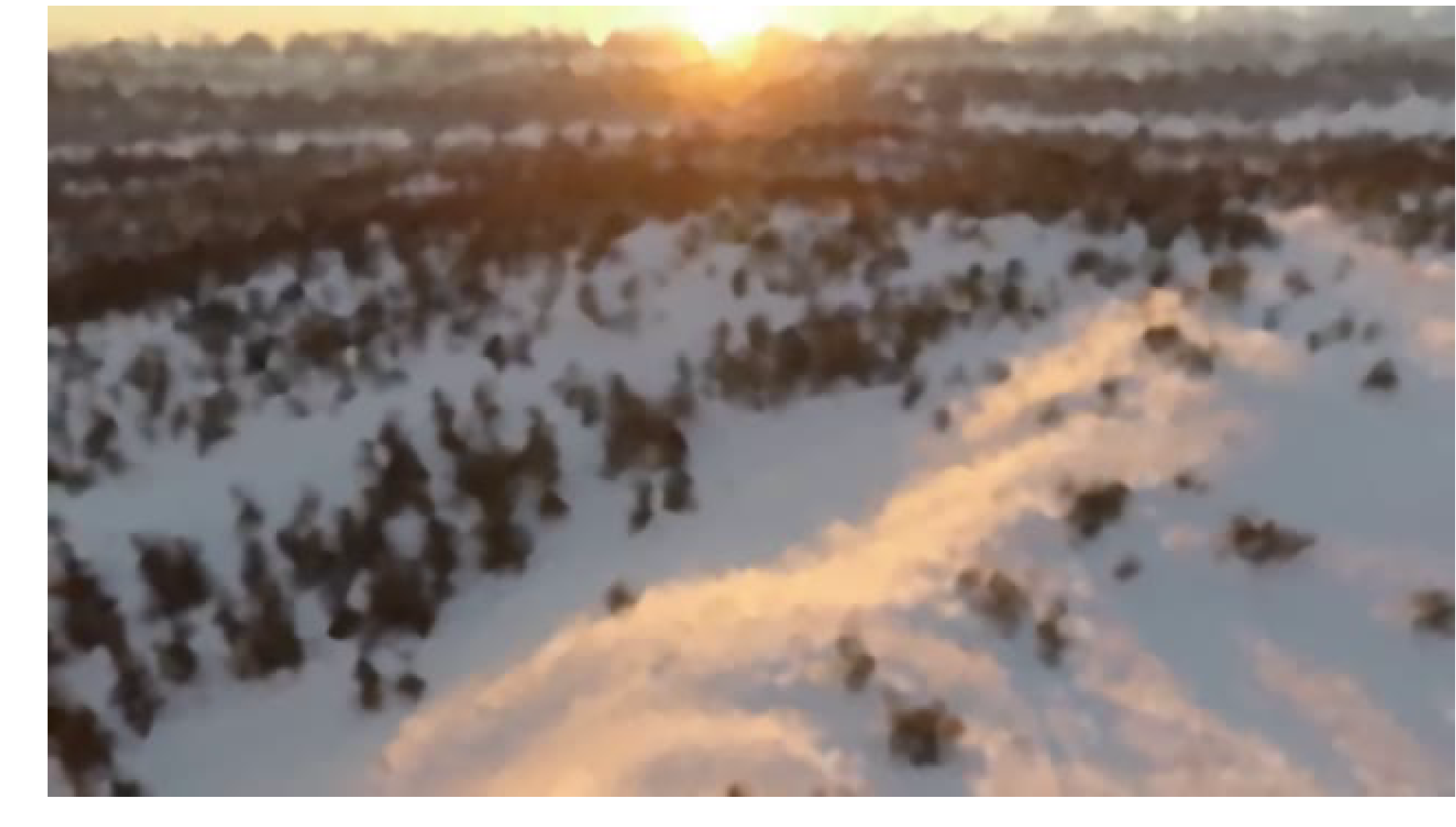}
	}
	\subfigure[EFAN3D (24.73 dB)]{
        \insetA{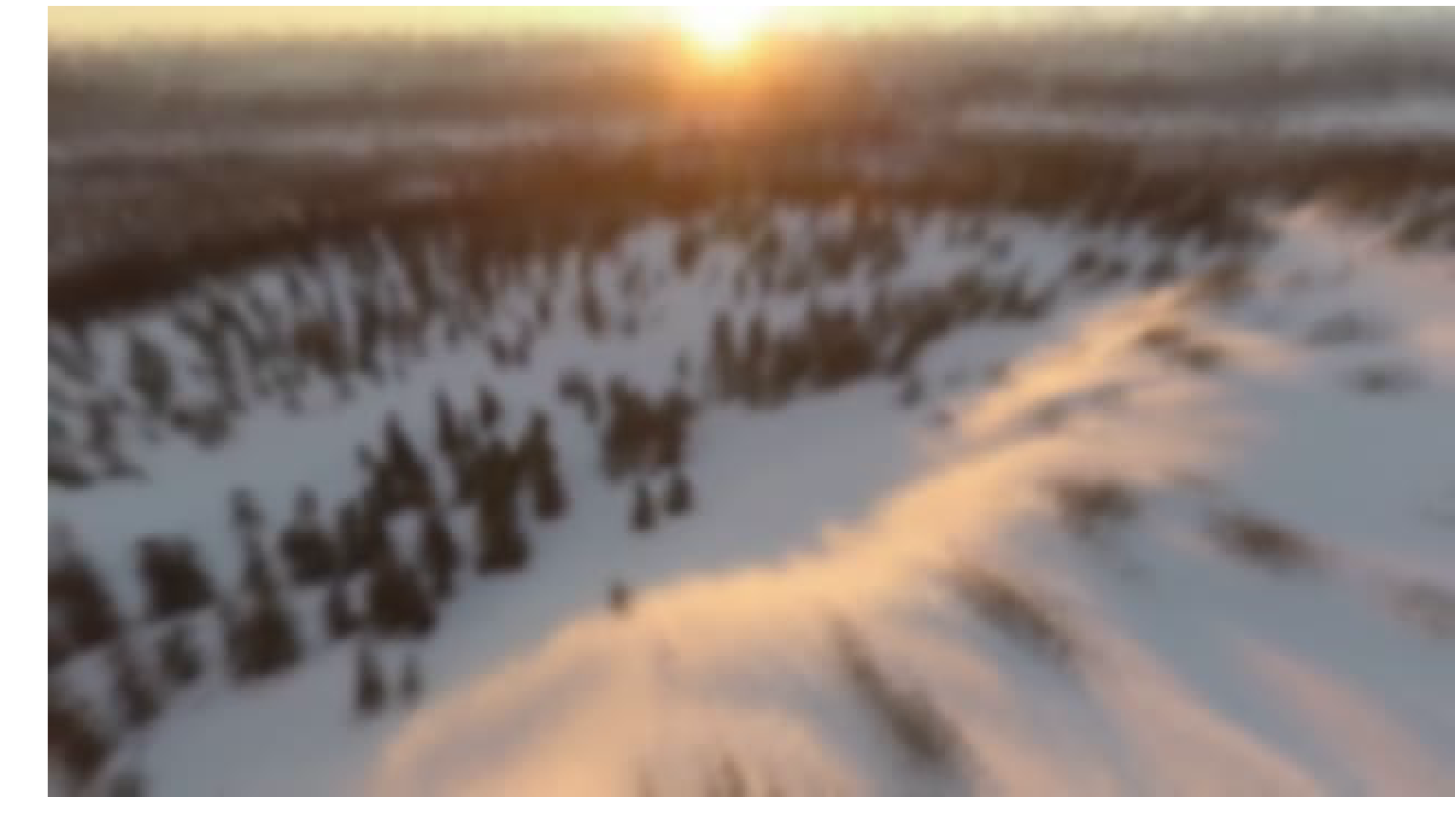}
	}
	\subfigure[ADEFAN (\textbf{26.03 dB})]{
        \insetA{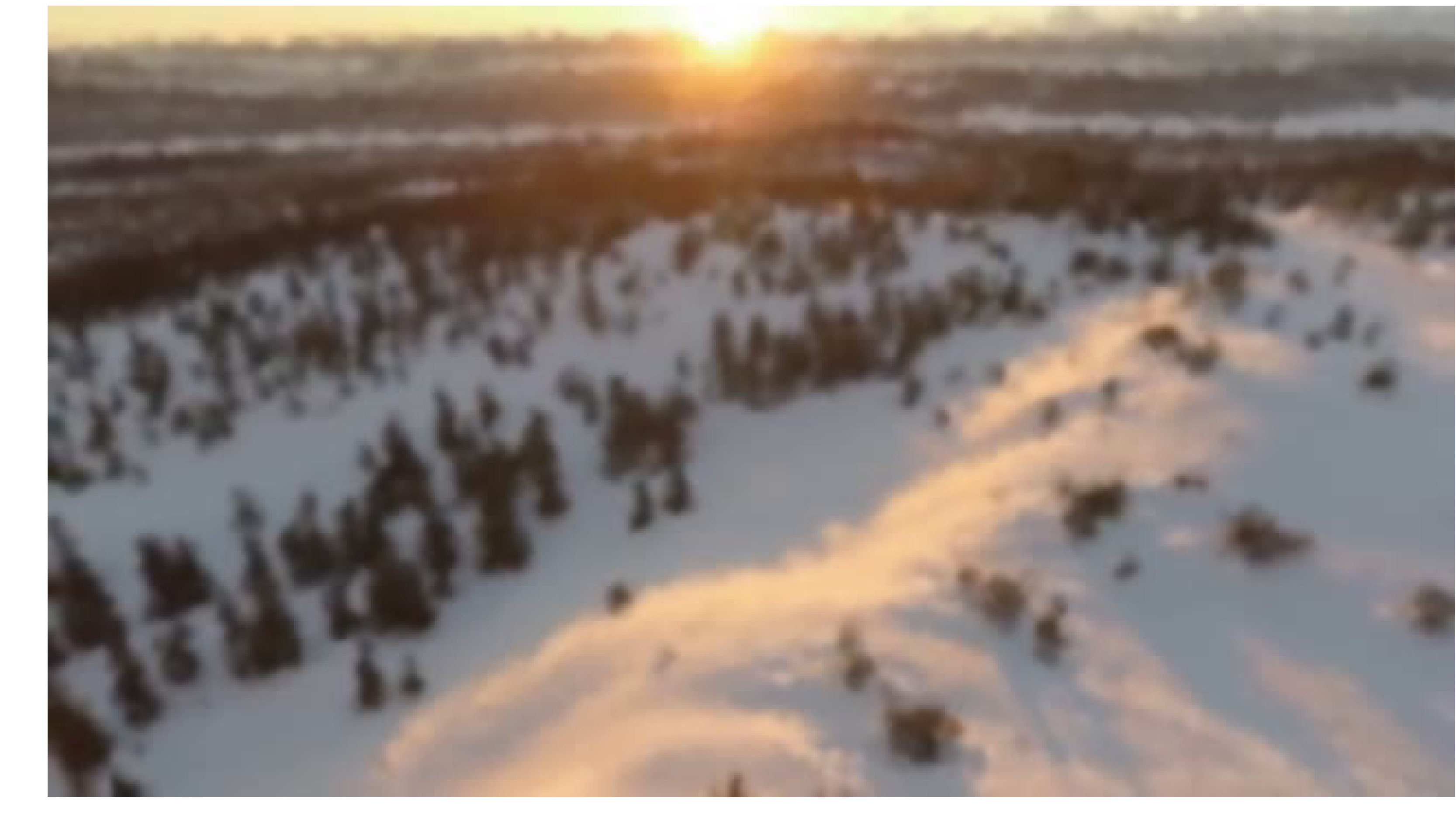}
	}
	\caption{The sample reference frame shown in (a) is reconstructed from 5\% random samples, using (b) EFAN2D, (c) EFAN3D and (d) our ADEFAN. While EFAN2D creates disk-like artifacts due to the sparsity of sampled pixels (visible in the magnified crop), EFAN3D causes excessive blurring. ADEFAN greatly reduces such artifacts and outperforms both methods in PSNR.}
	\label{fig:images_2}
\end{figure}

\section{Experimental Evaluation}


For all divergence estimation experiments, as well as for the evaluation of EFAN2D, EFAN3D and ADEFAN, we use a dataset consisting of $50$ videos with different content and optical flow velocities. All videos are publicly available\footnote{\label{footnote:dataset}\href{https://ieee-dataport.org/documents/extreme-video-completion-dataset}{https://ieee-dataport.org/documents/extreme-video-completion-dataset}}.

Our algorithms are implemented in Python 3 and will be made available\footnote{\href{https://github.com/majedelhelou/ADEFAN}{https://github.com/majedelhelou/ADEFAN}}. The evaluation is carried out on our full dataset of 50 videos. Fig.~\ref{fig:results_lineplot} shows the PSNR video reconstruction results of the different methods EFAN2D, EFAN3D and ADEFAN, for different sampling rates ranging from 1\% to 8\% of pixels. EFAN2D achieves good reconstruction PSNR at the expense of excessive flickering in the videos, which makes them disturbing to watch. On the contrary, EFAN3D smooths out the videos temporally, at the expense of reconstruction accuracy. Indeed, the PSNR results of EFAN3D are significantly worse than EFAN2D. Our proposed ADEFAN approach creates visually-pleasing reconstructions that do not suffer the flickering problem of EFAN2D, while also outperforming EFAN2D in terms of reconstruction PSNR (improving by $0.94dB$ on average for a sampling rate of 1\%). This is the case across all the sampling rates presented in Fig.~\ref{fig:results_lineplot}. We also show the PSNR results of all methods on every video in the dataset in Fig.~\ref{fig:results_barplot}. The only videos where EFAN2D outperforms ADEFAN in terms of PSNR are videos with extremely fast motion across the entire frames (e.g. Candle, Candle2).

We further validate our results with a survey. We collect 1,200 video ratings from 25 different participants on 8 randomly-selected videos\footref{footnote:dataset} and present the results in Table~\ref{tab:res_}. We ask for ratings on the reconstruction and the visual quality of the results. Users prefer the visual quality rather than the reconstruction quality of EFAN3D. The opposite is true for EFAN2D, which is nonetheless more preferred on average. ADEFAN is well-rated for both reconstruction and visual quality and is preferred over both competing methods.

\begin{figure}[t]
	\centering
	\subfigure[Reference video frame]{
		\includegraphics[width=0.42\linewidth]{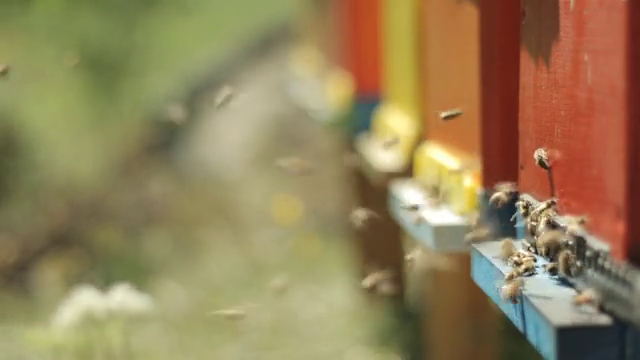}
	}
	\subfigure[ADEFAN (1\%) (26.53 dB)]{
		\includegraphics[width=0.42\linewidth]{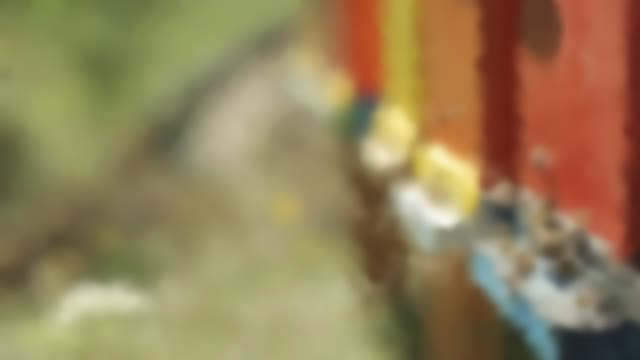}
	}
	\subfigure[ADEFAN (2\%) (\textbf{28.09 dB})]{
		\includegraphics[width=0.42\linewidth]{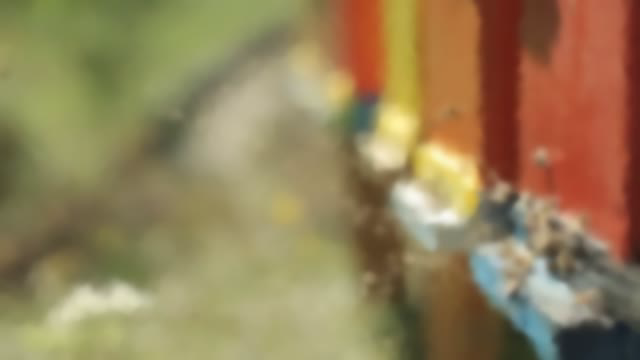}
	}
	\subfigure[MPEG-4 ($\approx$2\%) (26.03 dB)]{
		\includegraphics[width=0.42\linewidth]{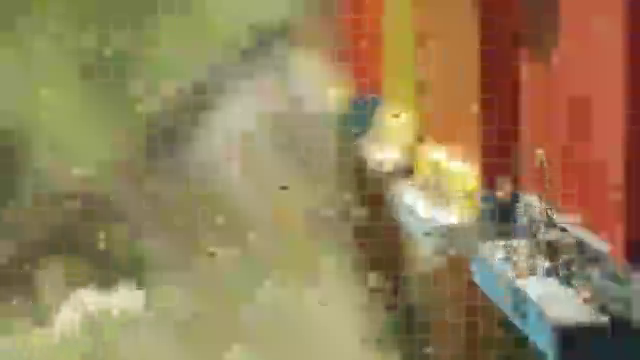}
	}
	\caption{The frame (a) is reconstructed with ADEFAN from 1\% and 2\% random samples and with MPEG-4 decompression from a slightly larger than 2\% MPEG-4 compression.}
	\label{fig:mpeg}
\end{figure}

For reference, we compare against the benchmark video compression algorithm MPEG-4~\cite{mpeg}, although this paper does not directly address compression. ADEFAN indeed has the limitation of having randomly-sampled pixels as input. For MPEG-4, we set the CRF quantizer scale to 51 to reach the highest compression rate~\cite{compress}. With the bitrate of the MPEG-compressed videos being slightly higher than the bitrate of the 2\%-sampled videos, the 1\%-sampled videos reconstructed using ADEFAN are of comparable quality, and those with 2\% of sampled pixels outperform MPEG-4. MPEG-4 compressed (down to 2\%) and decompressed videos have a PSNR of $25.35dB$ on our dataset, while ADEFAN achieves the higher PSNR results of $25.17dB$ and $26.54dB$ for the sampling rates of 1\% and 2\% respectively. In addition, videos reconstructed by ADEFAN are much smoother with no blocking artifacts. One such example is given in Fig.~\ref{fig:mpeg}. This shows the potential application of ADEFAN for extreme video compression.

\section{Conclusion} \label{sec:ccl}
We present a divergence-based adaptive extreme video completion approach that extends on the state-of-the-art efficient filtering by adaptive normalization~\cite{achanta}. It relies on color motion estimation based on KL divergence to adapt its temporal filtering and thus to leverage adjacent frames in the completion of any given video frame.

The proposed method can reconstruct extremely-sparsely sampled (e.g. 1\%) videos with randomly-sampled pixels. Our proposed completion method outperforms its competitors both in terms of reconstruction accuracy, assessed using PSNR and user opinions, and in terms of visual quality assessed through a survey.


\bibliographystyle{IEEE}
\bibliography{refs}

\end{document}